\documentclass[10pt,twocolumn,letterpaper]{article}

\usepackage[pagenumbers]{cvpr} 

\usepackage{graphicx}
\usepackage{amsmath}
\usepackage{amssymb}
\usepackage{booktabs}
\usepackage{microtype}
\usepackage{pifont}
\usepackage{color}
\usepackage{placeins}
\usepackage{float}
\usepackage{soul}
\usepackage{enumitem}
\usepackage{adjustbox}
\usepackage{commath}
\usepackage{bbm}
\usepackage{multirow}
\usepackage{caption}
\usepackage{array}
\usepackage{xcolor}
\usepackage{makecell}
\usepackage{subcaption}
\newcommand{\STAB}[1]{\begin{tabular}{@{}c@{}}#1\end{tabular}}

\definecolor{DarkGreen}{rgb}{0.0, 0.4, 0}
\definecolor{DarkYellow}{rgb}{0.4, 0.2, 0.0}
\definecolor{DarkPurple}{rgb}{0.44, 0.16, 0.39}
\definecolor{DarkRed}{rgb}{0.6,0,0}
\definecolor{DarkBlue}{rgb}{0,0,0.6}
\colorlet{LightYellow}{white!80!yellow}
\colorlet{LightRed}{white!80!red}
\colorlet{LightPurple}{white!80!purple}
\colorlet{LightBlue}{white!95!blue}
\colorlet{LightGreen}{white!80!green}

\DeclareRobustCommand{\hlyellow}[1]{{\sethlcolor{LightYellow}\hl{#1}}}

\DeclareRobustCommand{\hlblue}[1]{{\sethlcolor{LightBlue}\hl{#1}}}

\usepackage[pagebackref,breaklinks,colorlinks]{hyperref}

\usepackage[capitalize]{cleveref}
\crefname{section}{Sec.}{Secs.}
\Crefname{section}{Section}{Sections}
\Crefname{table}{Table}{Tables}
\crefname{table}{Tab.}{Tabs.}

\def\cvprPaperID{7939} 
\def\confName{CVPR}
\def\confYear{2022}

\begin{document}

\title{Induce, Edit, Retrieve: \\ Language Grounded Multimodal Schema for Instructional Video Retrieval}
\author{Yue Yang, Joongwon Kim, Artemis Panagopoulou, Mark Yatskar, Chris Callison-Burch\\
University of Pennsylvania\\
{\tt\small \{yueyang1, jkim0118, artemisp, myatskar, ccb\}@seas.upenn.edu}
}
\maketitle

\begin{abstract}
Schemata are structured representations of complex tasks that can aid artificial intelligence by allowing models to break down complex tasks into intermediate steps. We propose a novel system that induces schemata from web videos and generalizes them to capture unseen tasks with the goal of improving video retrieval performance. Our system proceeds in three major phases: (1) Given a task with related videos, we construct an initial schema for a task using a joint video-text model to match video segments with text representing steps from wikiHow; 
(2) We generalize schemata to unseen tasks by leveraging language models to edit the text within existing schemata. Through generalization, we can allow our schemata to cover a more extensive range of tasks with a small amount of learning data; 
(3) We conduct zero-shot instructional video retrieval with the unseen task names as the queries. Our schema-guided approach outperforms existing methods for video retrieval, and we demonstrate that the schemata induced by our system are better than those generated by other models.
\end{abstract}

\section{Introduction}
When encountering unfamiliar processes, people leverage knowledge from previous experience and generalize it to new situations. 
Cognitively, the information people use can be thought of as a \textit{schema}: a sequence of steps and a set of rules that a person uses to perform everyday tasks~\cite{widmayer2004schema}.
A schema can form a scaffold for adapting to unfamiliar contexts.
For example, a person may know the steps for baking a cake, and when confronted with a new task of baking a cupcake, she may try to modify a familiar cake process.
In this work, we study how a vision system can adopt such a reasoning approach and improve video retrieval.



\begin{figure}[!t]
\centering
    \includegraphics[width=8cm]{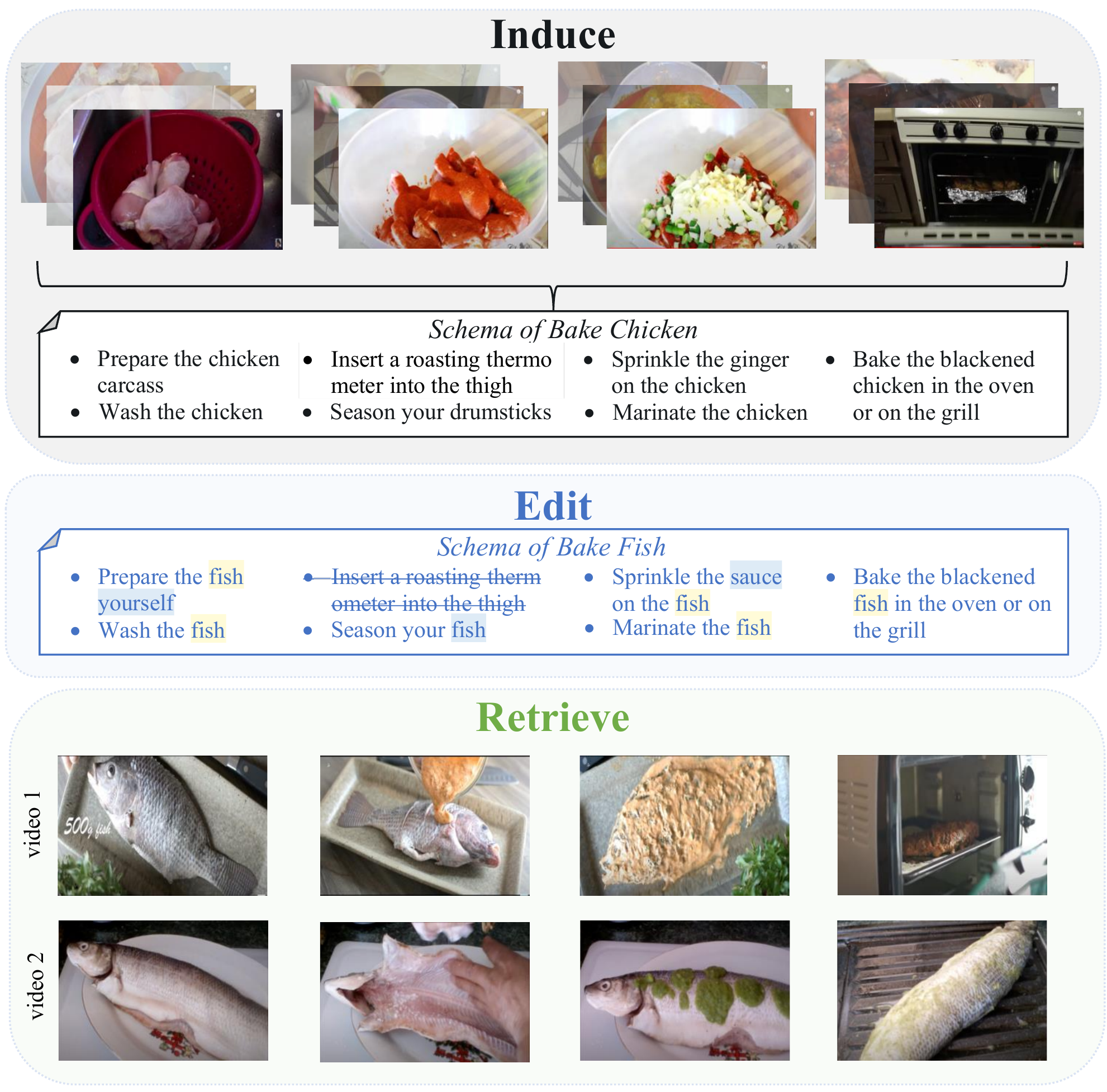}
    \caption{An example from our IER system, which first induces a schema for \textit{Bake Chicken} using a set of videos. Then it edits the steps in the schema to adapt to the unseen task \textit{Bake Fish} (the tokens that have been edited are \hlyellow{high}\hlblue{lighted}). Finally, IER relies on the edited schema to help retrieve videos for \textit{Bake Fish}.}
    \label{fig:pipeline_small}
\end{figure}


We propose a novel schema induction and generalization approach that we apply to video retrieval called \textbf{I}nduce, \textbf{E}dit, \textbf{R}etrieve (IER). 
Our schemata are represented as sets of natural language sentences describing steps associated with a task. 
Unlike pre-training approaches that construct implicit representations of procedural knowledge~\cite{zellers2021merlot}, our knowledge is explicit, interpretable, and easily adapted.
Furthermore, while others have tried to derive such knowledge directly from text~\cite{regneri2010learning, ostermann2020script, sakaguchi-etal-2021-proscript-partially}, we induce it from video. 
Once induced, our natural language schemata can be adapted to new unseen situations via explicit edit operations driven by BERT-based language models~\cite{devlin2018bert}. 
For example, IER is able to adapt an induced schema about \textit{Baking chicken} to a novel task of \textit{Baking fish}, as seen in Figure~\ref{fig:pipeline_small}.
Edited schemata can then be used to recognize novel situations and improve video retrieval systems.

We induce schemata by finding textual descriptions of videos that are reliably associated with a single task. 
Our system captions instructional YouTube videos from the Howto100M dataset~\cite{miech19howto100m} using candidate sentences from wikiHow~\footnote{\url{www.wikihow.com}}~\cite{lyu-zhang-wikihow:2020} with pretrained video-text matching models~\cite{miech20endtoend}. 
Sentences with the highest average matching score over all videos available for a task are retained for the schema.
The approach is simple and effective, leading to high-quality schemata with only 50 videos per task.
In total, we induce 22,000 schemata from 1 million videos with this approach. 

While large, our initial set of induced schemata is incomplete.
When faced with unseen tasks, we propose to adapt existing schemata using edit operations.
Our edits are directly applied to a schema's textual representation and are primarily guided by language models. 
Given a novel unseen target task, we pair it with a previously induced source task based on visual and textual similarity.
Then, we modify the steps in the source task's schema using three editing routines, as shown in Figure~\ref{fig:pipeline_small}.
Broadly, our edit operations first make object replacements to the schema using alignments between task names. 
For example, in Figure~,\ref{fig:pipeline_small} we change all instances of ``chicken'' to ``fish''.
Then we use a BERT~\cite{devlin2018bert} based model to both remove and modify the text in the source schema.
Sentences that are poorly associated with the target task name according to the model are removed.
Then we find low probability tokens and allow a BERT model to replace the ones with the lowest score with higher probability tokens~\cite{ghazvininejad2019mask}.
While this editing approach relies on finding sufficiently similar tasks in our induction set, our experiments show that our initial set of induced schemata can generalize to unseen tasks found in datasets such as COIN~\cite{tang2019coin} or Youcook2~\cite{ZhXuCoAAAI18}. 


The generated schemata can be used to retrieve multi-minute videos with extremely short queries~\footnote{On average, our queries are 4.4 tokens long.} in the form of task names.
Given a query, we retrieve videos using a schema from our initial induction set to produce a new schema through editing. 
The new schema is used to expand a short query into a larger set of sentences that can be matched to short clips throughout a long video. 
We evaluate the utility of our edited schemata for retrieval on Howto100M, COIN, and Youcook2 videos.
Results demonstrate that our IER approach is significantly better at retrieving videos than approaches that do not expand task names with schemata, improving nearly 10\% on top-1 retrieval precision.  
Furthermore, our edited schemata significantly outperform those generated from large language models such as GPT-3~\cite{brown2020language} in retrieval.
Finally, our extensive analysis shows that using schemata for retrieving instructional videos helps more as the length of the video increases.





\section{Related Work}
Previous work on \textit{schema induction} has focused solely on textual resources through statistical methods~\cite{chambers2009unsupervised, frermann2014hierarchical, belyy2020script,chambers2013event,pichotta2016statistical} and neural approaches~\cite{rudinger2015script,zhang-etal-2020-analogous,weber2018hierarchical,sakaguchi-etal-2021-proscript-partially,belyy2020script,Lyu-et-al:2021, li2020connecting,li2021future}. While~\cite{zellers2021merlot} employ multimodal resources to extract procedural knowledge, the output is an \textit{implicit} vector representation, unlike our work's explicit and interpretable schema. Another line of research~\cite{xu2020benchmark} extracts verb-arguments from video clips without aggregating information from multiple clips on the same topic. \cite{sener2019zero} aligns instructional text to videos in order to predict next steps but without schema generating method. To the best of our knowledge, this is the first attempt to extract explicit, human-readable schemata from videos and text. 

Prior work has followed the paradigm of template extraction and slot filling~\cite{kulkarni2013babytalk, lu2018neural, farhadi2010every,demirel2021detection, hou2019joint} for \textit{image/video captioning} to generalize to unseen situations and objects. While we draw inspiration from this literature, we instead retrieve human written sentences from wikiHow for captioning and employ language models to automatically modify them for unseen tasks. 


%


\textit{Graphical knowledge extraction} is not exclusive to script induction. A line of research that extracts graphical representations from visual input is scene graph extraction~\cite{zellers2018neural,yang2018graph,wang2019exploring,gu2019scene,gu2019scene, ji2020action}, i.e., the detection of objects and their relations from an image. Scene graphs have been applied to captioning~\cite{chen2020say,8630068,gu2019unpaired,yang2021reformer} and visual question answering~\cite{hudson2018compositional, hudson2019learning}.
While those methods rely on the same principle of extracting a graphical structure from visual input, the representations require explicit specification of label space for objects, attributes, and relations. 
In our work, instead, we let sentences stand in for structure. This allows us to leverage commonsense in language models to adapt our schemata.

For \textit{the text-video retrieval} task, earlier work has leveraged multimodal representations\cite{mithun2018learning,dong2019dual} to more effectively rank videos. Clip-based~\cite{gabeur2020multi,dzabraev2021mdmmt,wang2021t2vlad} and key-frame-based~\cite{dong2019dual,peng2004clip} ranking methods have been shown effective in improving retrieval performance. However, they rely on implicit multimodal representations rather than explicit, interpretable, and malleable representations as proposed in this work. Moreover, earlier methods generally focus on retrieving short video clips that are only several seconds long~\cite{ging2020coot, gabeur2020mmt}. While the videos in our retrieval task (see Table~\ref{tab: dataset stat}) can be multiple minutes long. 
\begin{figure*}[!t]
\centering
    \includegraphics[width=16cm]{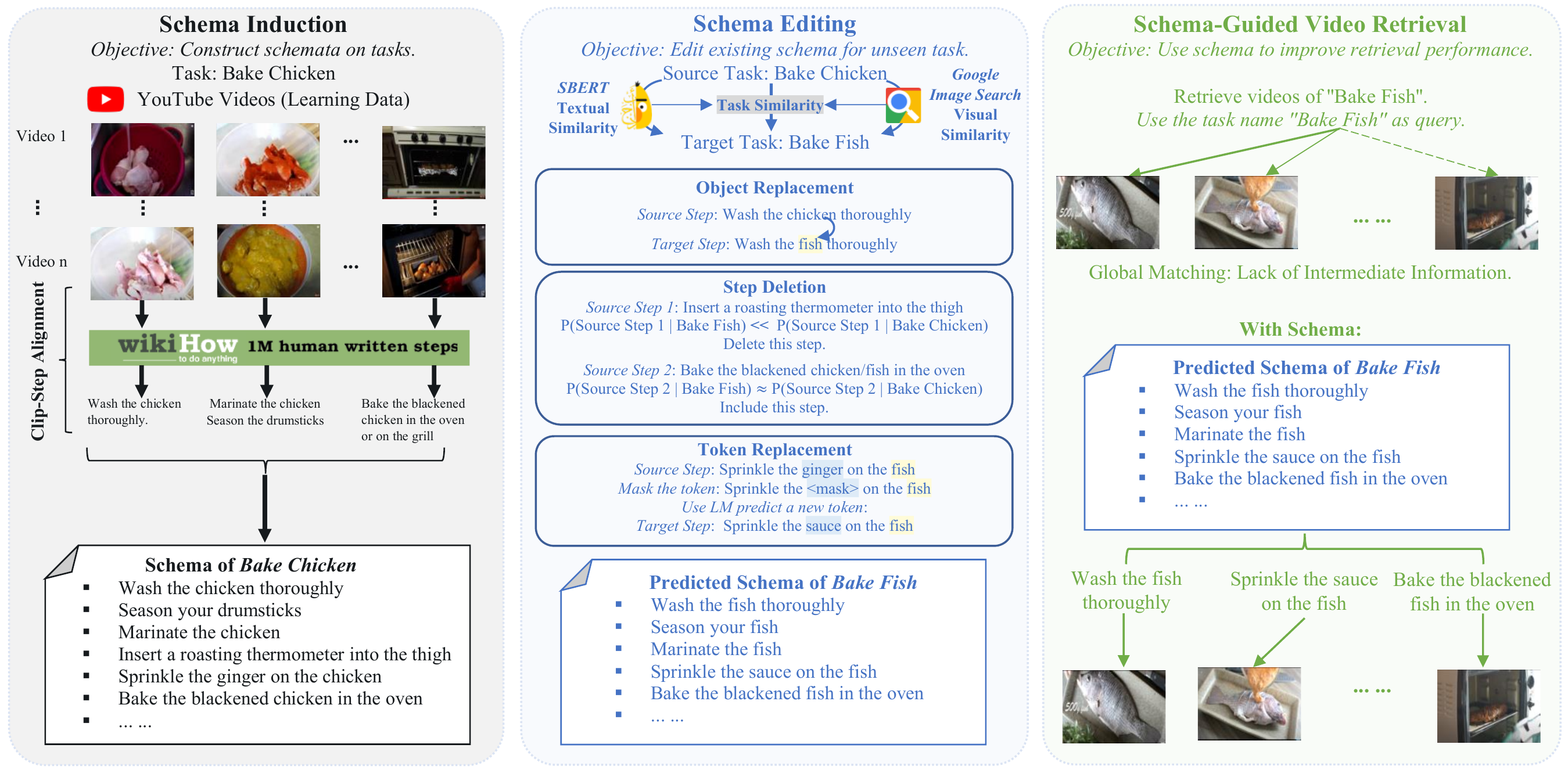}
    \caption{A detailed example of the IER system. The left panel demonstrates the induction phase, which takes in a set of videos describing the same task and outputs a schema in the format of a bag of sentences. The middle panel shows the schema editing system, which modifies the existing schema for unseen tasks, e.g., editing the schema of \textit{Bake Chicken} for the unseen task \textit{Bake Fish}. Finally, in the right panel, we use the edited schema of the unseen task to retrieve its associated videos by matching video segments with sentences in the edited schema.}
    \label{fig:pipeline large}
\end{figure*}

\section{Building a Schemata Library}
We create our schemata in two steps, shown in the first two panels of Figure~\ref{fig:pipeline large}:
(1) Schema induction, where schemata are generated for a set of tasks based on their associated videos, and (2) Schema editing, where schemata from the first phase are modified to address unseen tasks with no video data available.




\subsection{Formal Overview}
We assume a set of tasks $T$ partitioned into known tasks $K$ and unknown tasks $U$. Every task in the known set, $t \in K$, is associated with a set of videos $V_t$. We also assume a background textual corpus of candidate steps, $B$, made up of sentences describing tasks, not necessarily in $T$.

Our goal is to construct a schema, $S_t$, for every task  $t \in T$. We proceed in two steps. First, we use videos associated with tasks in $K$ to align sentences from $B$ using a matching function $F$ that scores pairs of short clips and sentences. The highest scoring alignments form the set of sentences in $S_t$. Second, given an unknown task, $t' \in U$, we find a similar source task $t \in K$ and modify its schema $S_{t}$ to create $S_{t'}$. 
\begin{table*}[!t]
\centering
\resizebox{17cm}{!}{%
\begin{tabular}{ccc}
\Xhline{2\arrayrulewidth}
\textbf{Object Replacement} & \textbf{Step Deletion} & \textbf{Token Replacement} \\ \hline
Cook Ham $\xrightarrow[]{\text{0.86}}$ Cook Lamb & Transplant a Young Tree $\xrightarrow[]{\text{0.89}}$ Remove a Tree & Prepare Fish $\xrightarrow[]{\text{0.82}}$ Prepare Crabs\\ 
Put the \textcolor{DarkYellow}{\hlyellow{ham}} in the oven. & Fill your pot with a balanced fertilizer. & Cut the \textcolor{DarkBlue}{\hlblue{fins}} from the \textcolor{DarkYellow}{\hlyellow{fish}} using \textcolor{DarkBlue}{\hlblue{kitchen}} \textcolor{DarkBlue}{\hlblue{shears}}.\\
$\downarrow$ & $\downarrow$\begin{small}delete\end{small}  & $\downarrow$\\
Put the \textcolor{DarkYellow}{\hlyellow{lamb}} in the oven. & \st{Fill your pot with a balanced fertilizer.} & Cut the \textcolor{DarkBlue}{\hlblue{shells}} from the \textcolor{DarkYellow}{\hlyellow{crabs}} using \textcolor{DarkBlue}{\hlblue{steel}} \textcolor{DarkBlue}{\hlblue{scissors}}.\\\hline
Clean a Guitar $\xrightarrow[]{\text{0.84}}$ Build a Violin & Fix a Toilet $\xrightarrow[]{\text{0.85}}$ Remove a Toilet & Make Healthy Donuts $\xrightarrow[]{\text{0.88}}$ Bake Healthy Cookies\\
Use a polish for particularly dirty \textcolor{DarkYellow}{\hlyellow{guitar}}s. & Test out the new flapper. & Slice your \textcolor{DarkYellow}{\hlyellow{donuts}} into \textcolor{DarkBlue}{\hlblue{disks}}.\\
$\downarrow$ & $\downarrow$\begin{small}delete\end{small}  & $\downarrow$\\
Use a polish for particularly dirty \textcolor{DarkYellow}{\hlyellow{violin}}s. & \st{Test out the new flapper.} & Slice your \textcolor{DarkYellow}{\hlyellow{cookies}} into \textcolor{DarkBlue}{\hlblue{squares}}.\\\hline
Trap a Rat $\xrightarrow[]{\text{0.84}}$ Trap a Rabbit & Brush a Cat $\xrightarrow[]{\text{0.87}}$ Brush a Long Haired Dog & Wash Your Bike $\xrightarrow[]{\text{0.84}}$ Wash a Motorcycle\\
Bait and set snap \textcolor{DarkYellow}{\hlyellow{rat}} traps. &  Comb and groom your pet. & Clean the \textcolor{DarkYellow}{\hlyellow{bike}} \textcolor{DarkBlue}{\hlblue{chain}} with a \textcolor{DarkBlue}{\hlblue{degreaser}}.\\
$\downarrow$ & $\downarrow$\begin{small}include\end{small}  & $\downarrow$\\
Bait and set snap \textcolor{DarkYellow}{\hlyellow{rabbit}} traps. & Comb and groom your pet. & Clean the \textcolor{DarkYellow}{\hlyellow{motorcycle}} \textcolor{DarkBlue}{\hlblue{thoroughly}} with a \textcolor{DarkBlue}{\hlblue{towel}}.\\ \Xhline{2\arrayrulewidth}
\end{tabular}
}
\caption{Examples of the operations performed by the three editing modules. Source task $\xrightarrow[]{G}$ Target task represents the generalization from a source task to a target task with task similarity score $G$, Source step $\downarrow$ Target Step denotes the editing of a source step to a target step. The \textcolor{DarkYellow}{\hlyellow{yellow}} words are replaced during the \textit{Object Replacement} operation and \textcolor{DarkBlue}{\hlblue{blue}} tokens are replaced by masked language model.}
\label{table:edit examples}
\end{table*}
\subsection{Schema Induction}\label{schema induction}
Given a known task $t \in K$, and its associated videos, $V_t$, we induce $S_t$ by retrieving sentences from $B$ that reliably describe steps performed in $V_t$.
Each video $v \in V_t$ can be partitioned into short segments or clips. For each segment $c \in v$, our goal is to find textual descriptions of the step being performed.

We use a pre-trained matching function $F$ between video and text to compute the matching score $F(c,s)$ between a segment $c$ and a step description $s$.
In practice, we use MIL-NCE~\cite{miech20endtoend}, a model trained on HowTo100M videos, to create video and textual embeddings with high similarity on co-occurring frames and transcripts.
For each clip $c$, we retain the 30 highest scoring step descriptions from $B$. 
Afterwards, for each step in the union of step descriptions retained for a task $t$, we average the matching score over all videos $V_t$ associated with the task:
\begin{equation}
    \text{score}(s, t) = \frac{1}{|V_t|} \sum_{v \in V_t} \Big( \frac{1}{|v|} \sum_{c \in v} F(c, s) \Big)
\end{equation}
We select the top-100 step descriptions for each task $t$ based on the score above.
Finally, we reduce redundancy by clustering similar descriptions.~\footnote{Paraphrases are very common in wikiHow, e.g., ``Remove the chicken from the oven'' and ``Remove your chicken from the oven'' both exist in corpus $B$. We use \href{https://scikit-learn.org/stable/modules/generated/sklearn.cluster.AgglomerativeClustering.html}{AgglomerativeClustering} API from sklearn for clustering.} 
We select the step with the highest matching score from each cluster to construct the schema $S_t$.~\footnote{On average, the number of sentences in each schema is 25.1.} 

\subsection{Schema Editing}\label{Schema Editing}
To produce the schema $S_{t'}$ for an unseen target task $t'$, we edit the schema of a similar source task $t$ in known set $K$. 
To achieve this objective, we develop a schema editing pipeline composed of three modules to manipulate the steps of the source schema (See Table \ref{table:edit examples} for examples).
Overall, our editing approach has three steps, performed in sequence, starting from deterministic replacements and ending with token-level edits performed by a language model. 
(1) {\bf Object Replacement}: we replace aligned objects from task names.
(2) {\bf Step deletion}: we remove irrelevant steps using a BERT-based question-answering system.
(3) {\bf Token Replacement}: we adjust steps at the token level by allowing a language model to replace tokens that have low probability. 
\medbreak
\noindent\textbf{Object Replacement} \hspace{0.2em}
Each task name has a main object, e.g., ``chicken'' in \textit{Bake Chicken}, found using a part-of-speech tagger.
For each task name, we retrain the first tagged noun as the main object.
We replace all occurrences of the main object in the source schema $S_t$ with the main object of the target task.
For example, in the first column of Table~\ref{table:edit examples}, we replace ``Ham'' with ``Lamb''.
\medbreak
\noindent\textbf{Step Deletion} Some steps are irrelevant for the new target task. 
For example, the task \textit{Bake Chicken} has a step ``Insert a roasting thermometer into the thigh'' which is inappropriate for the target task \textit{Bake Fish}.
Ideally, steps such as ``Preheat the oven'', which apply to both \textit{Bake Chicken} and \textit{Bake Fish}, will be preserved.

To identify which step to delete, we utilize a sentence BERT model~\cite{reimers-2019-sentence-bert} fine-tuned on question-answer pairs.\footnote{ \href{https://huggingface.co/sentence-transformers/multi-qa-mpnet-base-cos-v1}{\texttt{multi-qa-mpnet-base-cos-v1}}.} 
The model, $X$, computes a compatibility score between a question and an answer. 
It is trained to embed a question and an answer separately and then use the embedding similarity as the score. We use the model to score pairs of task names and steps and include a step in $S_{t'}$ when $X$ scores it as less compatible with $t'_q$ than $t_q$ by a significant margin: 
\begin{equation}\label{step deletion}
    \begin{cases}
        X(t'_q,s) < \beta \cdot X(t_q,s) & \text{Delete}  \\
        X(t'_q,s) \geq \beta \cdot X(t_q,s) &\text{Include}  \\
    \end{cases}
\end{equation}
where $\beta$ is a hyper-parameter determined on validation data. 
Examples of step deletions performed by our system can be found in the second column of Table~\ref{table:edit examples}.
\medbreak
\noindent\textbf{Token Replacement}
Finally, we adapt elements of the source task's schema at the token level, allowing a masked language model~\footnote{We choose \href{https://huggingface.co/distilroberta-base}{\texttt{distilroberta-base}}.} to replace words in a step with more appropriate alternatives.
We build on existing generation work using BERT-based models~\cite{ghazvininejad2019mask}.
We prompt the language model with a task name and a step, i.e., ``How to \underline{[TASK]}? \underline{[STEP]}'' and then greedily allow it to replace the least likely noun in the step with a higher scoring noun.
We repeat this iteratively on modified steps, a fixed number of times~\footnote{Determined by the number of nouns in a step.}.
For example, as in the third column of Table~\ref{table:edit examples}, we replace the word ``fins'' from a fish-based source task with ``shells'' in a crab-based target task.

\section{Schema Guided Video Retrieval} \label{retrieval methods}
To test the effectiveness of our schema induction and editing approaches, we formulate a novel video retrieval framework.
Given queries in the form of task names, we must retrieve long multi-minute videos corresponding to people instructing others on how to execute these tasks.
We use induced and edited schemata to retrieve such long videos.
We formulate a novel matching function that combines global information from the task name and steps information from the schema to retrieve such videos.
When using edited schemata, we average over multiple possible source tasks, allowing the model to combine information from multiple related tasks.




\subsection{Matching Function}
\medbreak
\noindent\textbf{Global Matching} 
Previous work on video retrieval largely focuses on short videos~\cite{ging2020coot, Luo2020UniVL, Luo2021CLIP4Clip}. 
They work predominately by matching a single feature vector, representing the entire video, to a query. 
However, in our retrieval scenario where videos are several minutes long, such an approach is impractical. 
Instead, given a query task, $t$, and a video, $v$ with associated segments $V_c$, we can average over a local matching score $F$, to estimate the overall compatibility between the task and the video:
\begin{equation}
    \mathcal{M}_{task}(t, v) = \frac{1}{|V_c|}\sum_{c \in V_c} F(c, t)
\end{equation}
This global averaging approach serves as the starting point for our schema-based retrieval function.

\medbreak
\noindent\textbf{Step Aggregation Model} Following~\cite{yang2021visual} who use sets of steps from wikiHow to match images, we define a video analog. 
The core idea is to score the compatibility between a schema, $S_t$, and a video, $v$, by finding an alignment between video segments and each sentence of the schema, $s \in S_t$. 
The alignment is done greedily, selecting the best video segment for each step in the schema. 
The average quality of these alignments can then be interpolated with the global score above, to form our final scoring function:

\begin{equation}\label{step agg}
    \begin{split}
        &\mathcal{M}_{step}(S_{t}, v) = \frac{1}{|S_{t}|}\sum_{s \in S_t} \max_{c \in V_c} F(c, s) \\
        &\mathcal{M}_{agg}(t, v) = (1 - \lambda)\mathcal{M}_{task}(t, v) + \lambda \mathcal{M}_{step}(S_t, v)
    \end{split}
\end{equation}
where $\lambda$ is a hyperparameter tuned on the development data.
Our scoring function smoothly interpolates between matching video directly with the task name and aligning video segments with the steps in the schema.  

\subsection{Task Similarity}
Our final retrieval system integrates over uncertainty in the schema. 
Since there are many possible source tasks for an unseen task, each of which can be used to predict a different schema, we average over possibilities. 
Each possibility is weighted by textual and visual similarity between the source and target task. 
This allows us to avoid using a schema from a task such as \textit{Bake Cake} to retrieve \textit{Bake Fish} videos. 

We score the similarity of tasks $t$ and $t'$ using textual ($G_{txt}$) and visual ($G_{vis}$), similarity between the two tasks:
\begin{equation}\label{eq:gen_score}
    G(t, t') = \max (G_{txt}(t, t'), G_{vis}(t, t'))
\end{equation}

\medbreak
\noindent\textbf{Textual Similarity} $G_{txt}$ is the sentence-level similarity computed by sentence-BERT \cite{reimers-2019-sentence-bert}.\footnote{We use \href{https://huggingface.co/sentence-transformers/all-mpnet-base-v2}{\texttt{all-mpnet-base-v2}} as the text encoder.} We compute the cosine similarity between the embeddings of $t$ and $t'$ extracted by sentence-BERT as $G_{txt}(t, t')$.

\medbreak
\noindent\textbf{Visual Similarity} $G_{vis}$ is computed from the image representations of the tasks.
For each task, we retrieve images from Google image search.~\footnote{We use \href{https://github.com/RiddlerQ/simple_image_download}{\texttt{simple\_image\_download}} package to get the urls of the Google images.}
Then we apply an image encoder\footnote{We use \href{https://github.com/openai/CLIP}{\texttt{clip-ViT-B-32}} as our image encodee.} to each image and average the resultant representations.
This aggregate vector is used to represent the visual embedding of the task.
We compute the cosine similarity between the features of source and target task as $G_{vis}(t, t')$.

\subsection{Video Retrieval on Unseen Tasks} 
In order to apply the step aggregation model in retrieving videos of an unseen task $t'$, we must find source task schemata to edit.
Given a target task $t'$, we first retrieve a set of $R$ most similar tasks, $T_s$, using $G$. 
For each retrieved source task $t_s \in T_s$, we construct an edited schema, $S_{t_s \rightarrow t'}$, using the routines defined in Section~\ref{Schema Editing}.
Edited schemata are integrated into retrieval based on task similarity, $G(t_s,t')$:
\begin{equation} \label{eq:agg with generalization}
\begin{split}
    \mathcal{M}(t', &v) = (1 - \lambda)\mathcal{M}_{task}(t', v) \\
    &+ \frac{\lambda}{R} \sum_{t_s \in T_s} G(t_s, t')\cdot\mathcal{M}_{step}(S_{t_s \rightarrow t'}, v)
\end{split}
\end{equation}

\section{Experiments}
This section will introduce the evaluation datasets and the baselines used for comparison, and the implementation details of our IER model.
\subsection{Datasets}
\noindent
\textbf{Howto100M} 
We use the Howto100M \cite{miech19howto100m} dataset for schema induction, as described in Section~\ref{schema induction}.
Howto100M is collected from YouTube using 1.22M instructional videos of 23k different visual tasks.
These visual tasks are selected from wikiHow articles, and each task is described by a set of step-by-step instructions in the article. The number of videos for each Howto100M task varies significantly. We keep the tasks that have at least 20 videos,  which results in 21,299 tasks.~\footnote{The videos of Howto100M are retrieved from Youtube, and each video is associated with a rank. We delete the videos with  Youtube search rank worse than 150 and assume these videos are not closely related to the task.} The task names are annotated by parts of speech (POS) in order to identify the main object for the Object Replacement operation during editing.~\footnote{We use the flair POS tagger \url{https://huggingface.co/flair/pos-english}.} 
\medbreak
\noindent
\textbf{Howto-GEN} To evaluate the schema editing modules, we split Howto100M tasks into two sets of known and unknown tasks. We select the tasks from Howto100M with exactly one noun, resulting in 3,365 tasks with 2,184 unique main objects. Then we randomly select 500 tasks for training and 500 tasks for validation and retrain 2,365 tasks for testing. Based on this split, there are 1,088 unseen main objects in the test set. We choose 5 videos for each test task for retrieval and pair them with a fixed set of 2,495 randomly sampled distractors videos to constitute a retrieval pool of 2,500 videos.~\footnote{We select the top-5 videos of each task for testing based on the Youtube search rank .}
\medbreak
\noindent
\textbf{COIN} \cite{tang2019coin} is a large-scale instruction video dataset with 11,827 videos for 180 tasks. The COIN tasks contain concepts unseen in Howto100M, such as ``Blow Sugar'', ``Play Curling'', ``Make Youtiao'', etc. We treat COIN as a zero-shot test set; we randomly pick five videos for every task~\footnote{No task names are shared between COIN and HowTo100M}. We finally construct a retrieval pool of 900 videos for 180 tasks.
\medbreak
\noindent
\textbf{Youcook2} \cite{ZhXuCoAAAI18} contains 2,000 long videos for 89 cooking recipes. We treat recipe names as tasks and use the same split as \cite{miech19howto100m} to guarantee that there is no overlap between the videos in Youcook2 and Howto100M. We finally form a retrieval pool of 436 videos.
\begin{table}[!t]
\centering
\resizebox{8cm}{!}{%
\begin{tabular}{cccc}
\Xhline{2\arrayrulewidth}
\textbf{Dataset}   & \textbf{\# of tasks} & \textbf{\# of videos} & \textbf{Avg. video length (s)} \\ \hline
Howto-GEN & 2,365      & 11,825       &   392.9                       \\
COIN      & 180        & 900          &   143.2                       \\
Youcook2  & 89         & 436          &   310.9\\ \Xhline{2\arrayrulewidth}                      
\end{tabular}
}
\caption{Statistics of the evaluation datasets (test set).}
\label{tab: dataset stat}
\end{table}

\subsection{Preprocessing}
The video segments boundaries in Howto100M are generated from an Automatic Speech Recognition system and are noisy and redundant.
To reduce the number of segments per video, we apply \textit{k}-means to the S3D features \cite{xie2018rethinking} of the clips, iteratively range \textit{k} from 5 to 10 and select the best \textit{k} with the highest silhouette score \cite{ROUSSEEUW198753}.
Then we pick the segment nearest to the center of each cluster to form the sequence of clips for each video.
For COIN and Youcook2, we use the human-annotated video segments provided in their dataset.

\subsection{Baselines}
\noindent
\textbf{Global Matching} We leverage the MIL-NCE model with the global averaging method described in Section \ref{retrieval methods} to retrieve procedural videos.
\medbreak
\noindent
\textbf{Step Aggregation Model} As proposed in Section \ref{retrieval methods}, we use edited schemata to improve video retrieval performance. 
For comparison, we use alternative methods to expand task names into schemata:
\begin{itemize}[leftmargin=*]
\itemsep0em
    \item \textbf{T5}~\cite{Lyu-et-al:2021} We propose a generation-based schema induction approach and fine-tune a multilingual T5 model \cite{xue-etal-2021-mt5} using the wikiHow scripts. The model can generate a list of steps given a task as the prompt.
    \item \textbf{GPT-2}~\cite{radford2019language} Following the same experimental setup as T5, we fine-tune a GPT-2-large model to generate the schemata.
    \item \textbf{GPT-3}~\cite{brown2020language} We use the OpenAI GPT-3 (davinci) model to conduct zero-shot schema generation using the prompt - ``How to \underline{Task Name}? Give me several steps.''.
    \item \textbf{GOSC} Goal-Oriented Script Construction (GOSC) \cite{Lyu-et-al:2021} is a retrieval-based approach to construct a schema. GOSC utilizes a Step Inference model to gather the set of desired steps from wikiHow given the input task name. We use the off-the-shelf model, so some of the Howto-GEN test tasks have been seen during the training process of GOSC.
    \item \textbf{wikiHow} We treat wikiHow as a schema library. For each unseen test task, we find the most similar task in wikiHow based on the similarity score and apply the schema editing modules to obtain the edited schema.
    \item \textbf{Oracle} Our oracle schemata are written by humans for all datasets. For Howto-GEN, the oracle schemata are the steps in the exact, corresponding wikiHow articles. COIN provides human-annotated step labels for each task which we consider as the oracle schemata. For Youcook2, we treat the text annotations of the video segments as the oracle schemata.
\end{itemize}
\begin{table*}[!t]
\centering
\resizebox{17cm}{!}{%
\begin{tabular}{cc|ccccc|ccccc|ccccc}
\Xhline{2\arrayrulewidth}
\multicolumn{2}{c|}{\multirow{2}{*}{\textbf{Method}}} & \multicolumn{5}{c|}{\textbf{Howto-GEN}} & \multicolumn{5}{c|}{\textbf{COIN}} & \multicolumn{5}{c}{\textbf{Youcook2}}            \\ \cline{3-17}
\multicolumn{2}{c|}{}  & \textbf{P@1}$\uparrow$ & \textbf{R@5}$\uparrow$ & \textbf{R@10}$\uparrow$ & \textbf{Med r}$\downarrow$ & \textbf{MRR}$\uparrow$ & \textbf{P@1}$\uparrow$ & \textbf{R@5}$\uparrow$ & \textbf{R@10}$\uparrow$ & \textbf{Med r}$\downarrow$ & \textbf{MRR}$\uparrow$ & \textbf{P@1}$\uparrow$ & \textbf{R@5}$\uparrow$ & \textbf{R@10}$\uparrow$ & \textbf{Med r}$\downarrow$ & \textbf{MRR}$\uparrow$\\ \hline
\multirow{1}{*}{} & MIL-NCE~\cite{miech20endtoend}                                                & 45.2 & 31.0 & 43.1 & 15.0 & .198 & 48.3 & 37.1 & 52.8  & 9.5 & .227 & 27.0 & 18.2 & 26.5 & 32.0 & .126 \\ \hline
\multirow{7}{*}{\STAB{\rotatebox[origin=c]{90}{Step Aggregation}}}         & T5~\cite{Lyu-et-al:2021} & 44.0 & 29.9 & 41.0 & 19.0 & .190 & 46.1 & 35.3 & 50.7 & 10.0 & .219 & 21.3 & 16.0 & 24.7  & 61.5 & .108 \\
                                                                           & GPT-2~\cite{radford2019language} & 46.0 & 31.5 & 43.3 & 16.0 & .200 & 48.9 & 39.2 & 53.4 & 8.0 & .233 & 31.5 & 19.0 & 27.3 & 44.5 & .130 \\
                                                                           & GPT-3~\cite{brown2020language} & 49.3 & 33.3 & 45.7 & 13.0 & .211 & 53.3 & 42.1 & 59.0 & 8.0 & .252 & 37.1 & 22.4 & 34.6 & 27.0 & .160 \\
                                                                           & GOSC~\cite{Lyu-et-al:2021} & 54.7 & 37.0 & 49.8 & 11.0 & .231 & 53.9 & 41.6 & 55.1 & 8.0 & .248 & 30.3  & 20.7 & 34.8 & 28.0 & .146    \\
                                                                           & wikiHow & 51.9 & 35.4 & 47.8 & 11.0 & .222 & 53.9 & 40.8 & 56.1 & 7.0 & .246 & 31.5 & 21.0 & 34.2  & 24.5 & .149 \\
                                                                           & IER (Ours) & 54.4 & 37.3 & 50.1 & 10.0 & .231& \textbf{57.2} & 42.2 & 57.8 & \textbf{7.0} & .256 & \textbf{41.6} & \textbf{25.8} & \textbf{38.8} & \textbf{20.0} & \textbf{.175}\\
                                                                           & IER$^3$ (Ours) & \textbf{55.0} & \textbf{37.4} & \textbf{50.6} & \textbf{10.0} & \textbf{.234} & 56.1 & \textbf{42.3} & \textbf{59.1} & 8.0 & \textbf{.258} & 40.4 & 25.1 & \textbf{38.8} & \textbf{20.0} & .172\\ \cline{2-17}
                                                                         
                                                                           & Oracle       & 56.5 & 38.0 & 50.8 & 10.0 & .237 & 60.0 & 43.4 & 59.3 & 7.0 & .262 & 52.8 & 33.5 & 47.1 & 14.0 & .215    \\ \Xhline{2\arrayrulewidth}
\end{tabular}
}
\caption{Retrieval performance on Howto-GEN, COIN and Youcook2. Baselines include retrievals based on global matching, aggregation of steps generated from state-of-the-art language models, goal-oriented script construction (GOSC), and wikiHow. The Oracle upper bound contains human-written step labels for each task. Observe that our \textbf{IER} systems outperform the baselines across all metrics.}
\label{main_results}
\end{table*}

\subsection{Implementation Details}
\medbreak
\noindent
\textbf{Hyperparameters} We fine-tune the hyperparameters on the validation set of Howto-GEN. We set $\beta = 0.8$ in equation \ref{step deletion} as the threshold to determine which step to remove. We select $\lambda = 0.6$ to adapt the weight of the step score in equation \ref{step agg}. The two hyperparameters are fixed for all tests. 
\medbreak
\noindent
\textbf{IER} When evaluated on the Howto-Gen test set, the IER model can only have access to the schemata of 500 training tasks. Meanwhile, for COIN and Youcook2, IER can use all 21,299 schemata learned from Howto100M. As described in equation \ref{eq:agg with generalization}, we can select multiple schemata to assist retrieval. We report the performance of IER with the top-1 schema and the top-3 schemata (IER$^3$) in the results.

\subsection{Evaluation Metrics}
We use the standard metrics to evaluate retrieval performance: Precision@1 (P@1), Recall@K (R@K), Mean rank (Mean r), Median rank (Med r), and Mean Reciprocal Rank (MRR). We use $\uparrow$ or $\downarrow$ to indicate whether a higher or lower score is better in all tables and figures.

\begin{figure}[!t]
\centering
    \includegraphics[width=7cm]{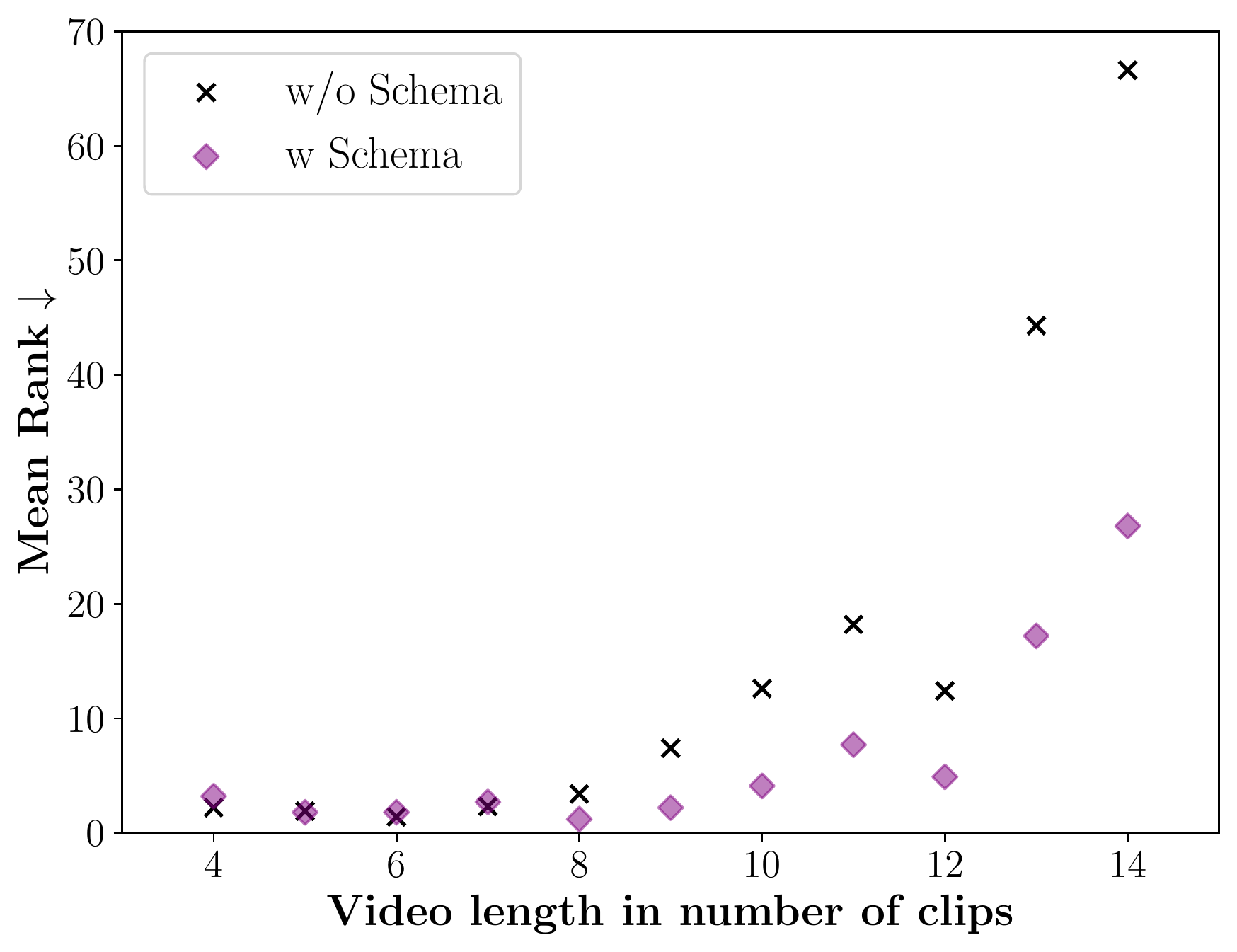}
    \caption{Retrieval performance by video length (in the number of clips). We group the test videos of Youcook2 by the number of clips per video and compute the mean rank for each group.}
    \label{fig:long video}
\end{figure}

\section{Results}
\subsection{Main Results}
As shown in Table \ref{main_results}, almost all step aggregation models assisted with schemata outperform the MIL-NCE model except for T5. These results suggest that the use of schemata is a promising way to enhance the retrieval of procedural videos. Furthermore, our IER model outperforms the other purely textual schema induction baselines and is close to the performance of the oracle. 

We analyze the retrieval performance by video length in Figure \ref{fig:long video}. The performance of the model without schemata declines rapidly as video length increases. However, when using schemata induced and edited by IER, the performance declines substantially less on long videos.

\begin{figure}[!t]
\centering
    \includegraphics[width=7cm]{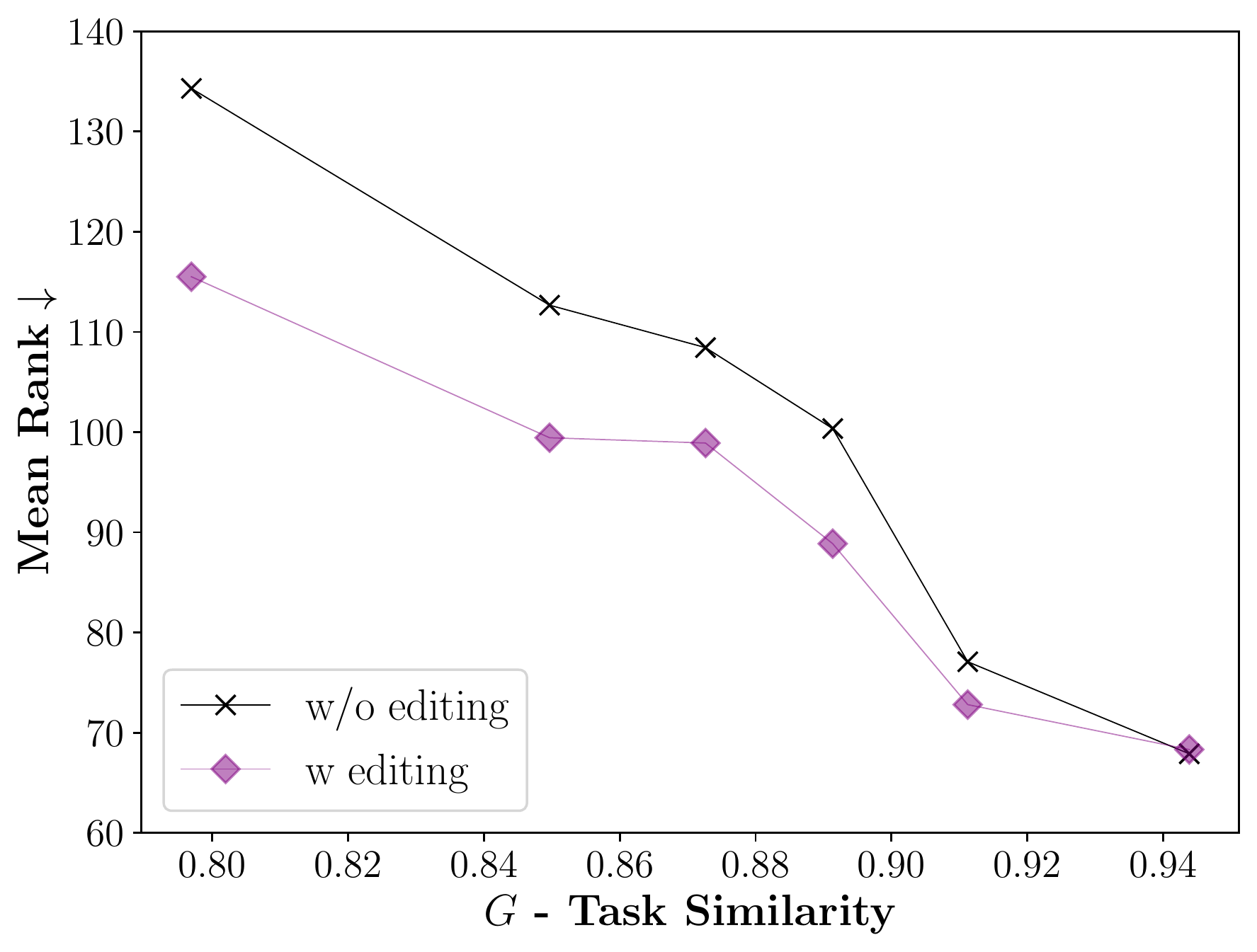}
    \caption{Retrieval performance by task similarity. We sort the test tasks of Howto-GEN based on their task similarity ($G$) and compute their mean rank for every batch of 400 tasks.}
    \label{fig:generalization impact}
\end{figure}


\begin{figure*}
      \centering
\begin{subfigure}[b]{0.48\textwidth}
\includegraphics[width=\textwidth]{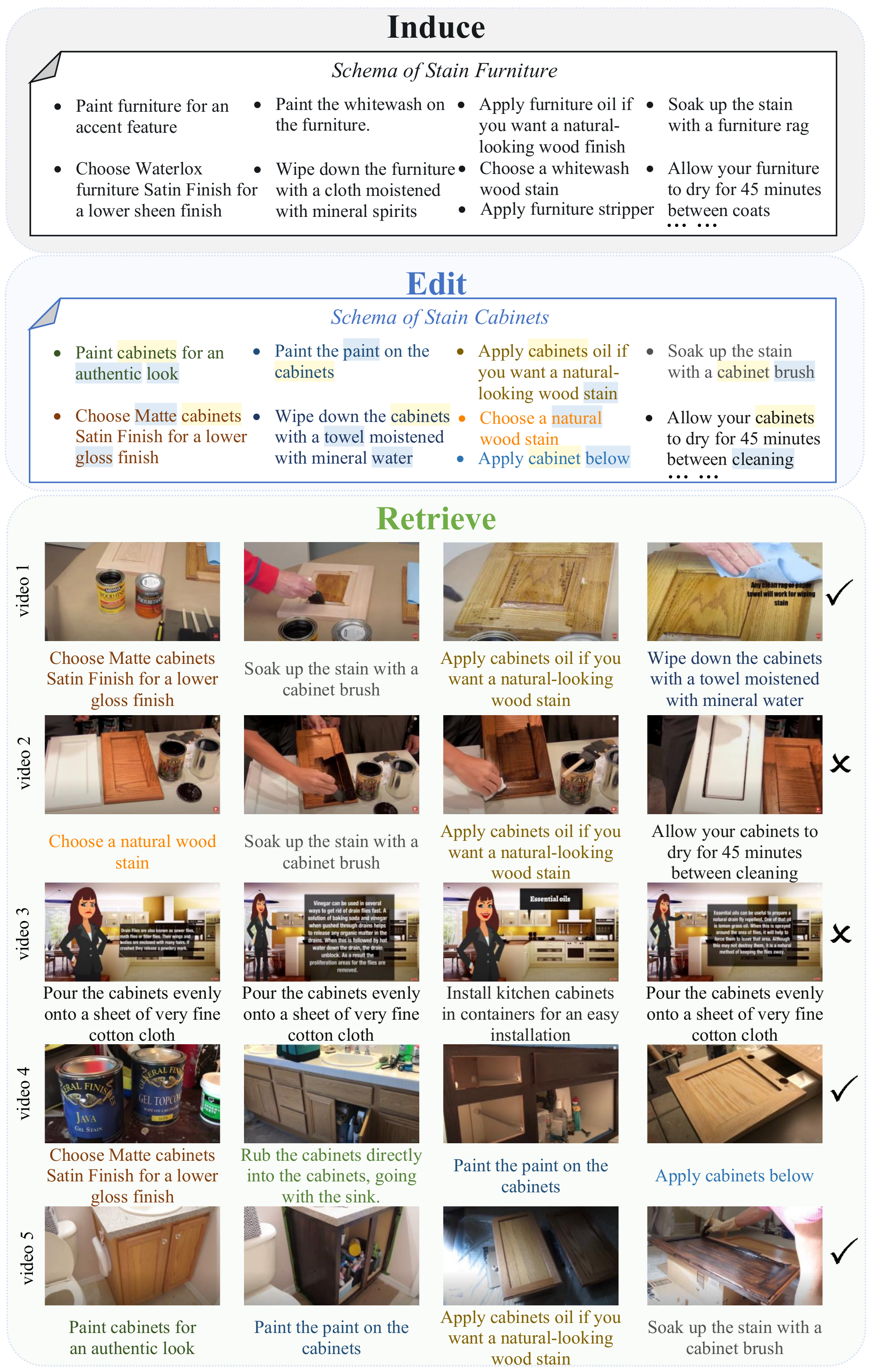}
\caption{Stain Cabinets}
\end{subfigure}
\begin{subfigure}[b]{0.48\textwidth}
\includegraphics[width=\textwidth]{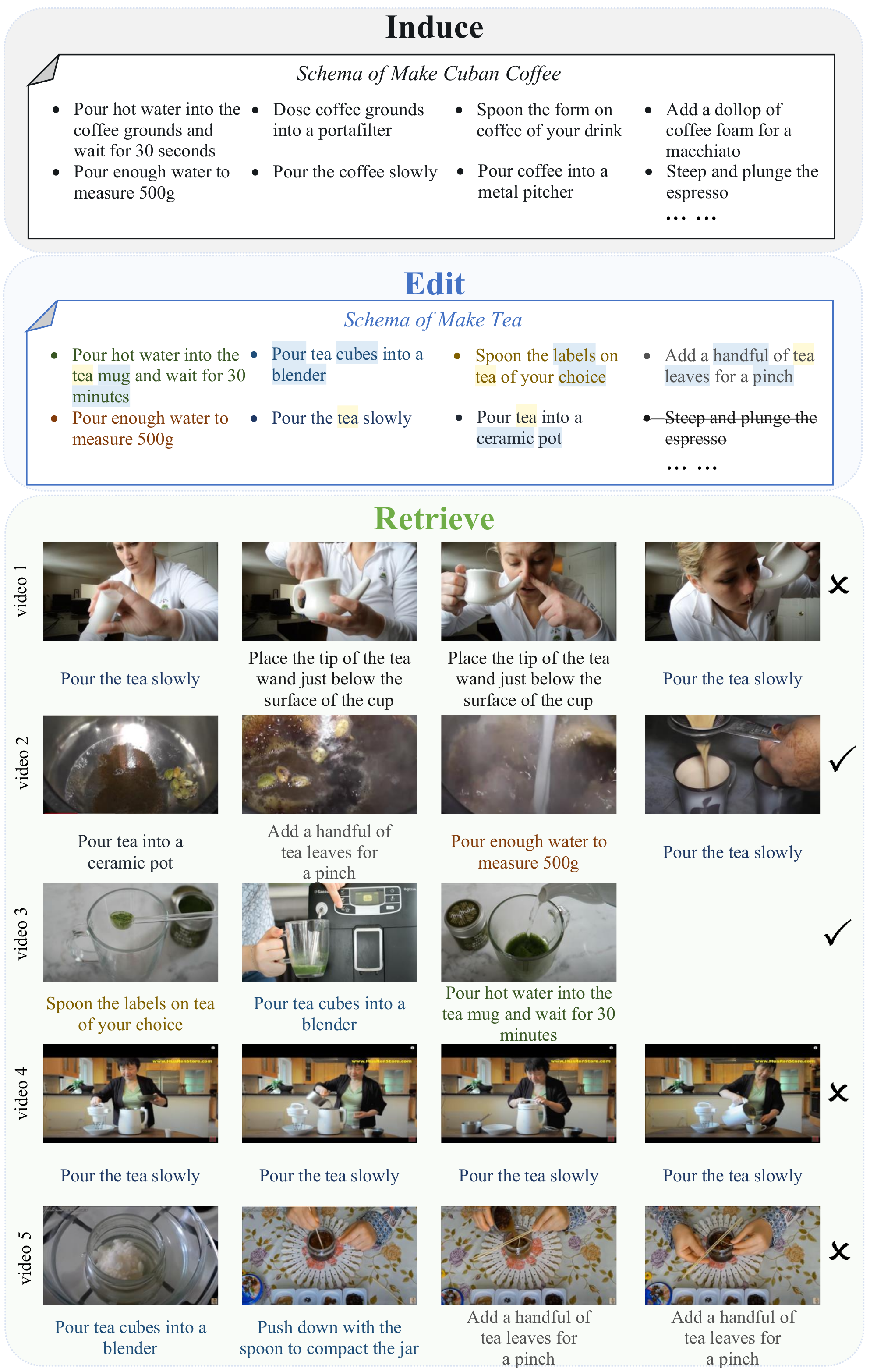}
\caption{Make Tea}
\end{subfigure}
%
\caption{Qualitative examples of retrieval results. We demonstrate our video retrieval process for two unseen tasks. On the top, we display the induced schemata of existing tasks. Below, we display schemata for the unseen tasks obtained from our editing module. Finally, we display the top-5 videos retrieved using our edited schemata. We show at most 4 segments for each video and each segment is associated with the top-1 matched step in the schema. While many of our videos are correctly retrieved, some videos are not due to two main factors: 1) our joint video-text model propagates some errors, and 2) some of the videos are labeled under different tasks while containing very similar steps, such as video 2 which belongs to the task \textit{Apply a Two Tone Finish to Furniture} that is very close to the query \textit{Stain Cabinets}.}
\end{figure*}

\begin{table}[!t]
\centering
\resizebox{8cm}{!}{%
\begin{tabular}{c|c|ccccc}
\Xhline{2\arrayrulewidth}
                    & \textbf{Method}         & \textbf{P@1}$\uparrow$ & \textbf{R@5}$\uparrow$ & \textbf{R@10}$\uparrow$  & \textbf{Med r}$\downarrow$ & \textbf{MRR}$\uparrow$ \\ \hline
\multirow{5}{*}{\STAB{\rotatebox[origin=c]{90}{\textbf{Howto-GEN}}}} & full   &  \textbf{54.4}   &  \textbf{37.3}   &  \textbf{50.1}     &  \textbf{10.0}     &  \textbf{.231}   \\ \cline{2-7} 
                           & $-$ mask        &  \underline{53.7}   & \underline{36.3}   &  \underline{49.3}    &  \underline{11.0}      & \underline{.229}    \\
                           & $-$ deletion    &  \underline{53.6}   & \underline{36.9}   &  \underline{49.8}    &  \underline{11.0}      & \underline{.230}    \\
                           & $-$ replacement &  \underline{51.5}   & \underline{34.9}   &  \underline{47.3}    &  \underline{12.0}      & \underline{.220}    \\
                           & $-$ all         &  \underline{45.5}   & \underline{31.0}   &  \underline{43.1}    &  \underline{15.0}      & \underline{.199}    \\ \hline \hline
\multirow{5}{*}{\STAB{\rotatebox[origin=c]{90}{\textbf{COIN}}}} & full   &  57.2   &  42.2   &  57.8    &  \textbf{7.0}    &  .256   \\ \cline{2-7} 
                           & $-$ mask        &  \underline{53.9}   &  \textbf{42.3}   &  58.3    &  \textbf{7.0}      &  .257    \\ 
                           & $-$ deletion    &  \textbf{58.3}   &  \underline{42.0}   &  58.0    &   \textbf{7.0}     &  \textbf{.258}   \\
                           & $-$ replacement &  \underline{53.8}   &  \underline{41.0}   &  \textbf{59.2}    &   \underline{7.5}     &  \underline{.251}   \\
                           & $-$ all         &  \underline{54.4}   &  \underline{39.6}   &  \underline{53.7}    &   \underline{8.0}     &  \underline{.246}   \\ \hline \hline
\multirow{5}{*}{\STAB{\rotatebox[origin=c]{90}{\textbf{Youcook2}}}} & full   &  \textbf{41.6}   & 25.8    & 38.8     & \textbf{20.0}    & \textbf{.175}  \\ \cline{2-7} 
                           & $-$ mask        & \underline{40.4}    & \underline{25.4}    & 39.3    & \textbf{20.0}       & \underline{.173}    \\ 
                           & $-$ deletion    & \textbf{41.6}    & \textbf{26.0}    & 39.1     & \underline{21.0}       & \textbf{.175}    \\
                           & $-$ replacement & \underline{40.4}    & 25.8    & \underline{38.5}     & \textbf{20.0}       & \underline{.173}    \\ 
                           & $-$ all         & \underline{40.4}    & \textbf{26.0}    & \textbf{39.9}     &  \underline{21.0}      & \underline{.174}    \\ \Xhline{2\arrayrulewidth}
\end{tabular}
}
\caption{Ablation study on editing modules. ``full'' represents using all three modules and ``$-$ all'' denotes removing all three modules. ``$-$ mask'', ``$-$ deletion'' and ``$-$ replacement'' are short for removing ``Token Replacement'', ``Step Deletion'' and ``Object Replacement'' respectively. The numbers with \underline{underline} are the ones lower than ``full''. The highest number of each metric is \textbf{bold}.}
\label{table:gen ablation}
\end{table}
\subsection{Editing Module Ablations}

To validate whether each editing module benefits the retrieval, we conduct an ablation study where we disable these modules one by one. As shown in Table~\ref{table:gen ablation}, the editing modules never hurt and often improve retrieval performance for Howto-GEN and COIN. However, the editing modules are not necessary for Youcook2 because the tasks of Youcook2 are very close to the ones in Howto100M, and we can always find schemata of similar tasks. As shown in Figure \ref{fig:generalization impact}, editing is more useful when task similarity is low.~\footnote{We compute the average task similarity for each dataset, Howto-GEN is 0.88, COIN is 0.92, and Youcook2 is 0.97, which explains why editing modules are not helpful for Youcook2.} 

\subsection{Schemata Transfer}
Our schemata can improve video retrieval even when used with representations they were not induced on.
For example, we experiment with CLIP\cite{radford2021learning}. 
Following \cite{portillo2021straightforward, Luo2021CLIP4Clip}, which leverage CLIP for video via average-pooling, we convert video clips into sequences sampled at 10 FPS.
Then we use \texttt{clip-ViT-B-32} to encode each frame and average over the frame-level features for video representations. This allows us to use CLIP as the matching function $F$.

\begin{table}[!t]
\centering
\resizebox{8cm}{!}{%
\begin{tabular}{ccccccc}
\Xhline{2\arrayrulewidth}
                    & \textbf{Model}         & \textbf{P@1}$\uparrow$ & \textbf{R@5}$\uparrow$ & \textbf{R@10}$\uparrow$  & \textbf{Med r}$\downarrow$ & \textbf{MRR}$\uparrow$ \\ \hline
                    & MIL-NCE & 48.3 & 37.1 & 52.8 & 9.5 & .227\\
                    & $+$schema & 57.2 & 42.2 & 57.8 & 7.0 & .256\\ \hline
                    & CLIP\cite{radford2021learning} & 58.9 & 44.9 & 58.8 & 6.0 & .264\\
                    & $+$schema & 65.0 & 47.4 & 60.8 & 5.5 & .282\\ \Xhline{2\arrayrulewidth}
\end{tabular}
}
\caption{Retrieval performance on COIN using MIL-NCE and CLIP as the matching functions. $+$schema represents using schema induced by IER (MIL-NCE as matching function) for retrieval.}
\label{table: clip}
\end{table}

We compute the retrieval performance of CLIP on COIN using the global matching method and the step aggregation method with the same schemata as MIL-NCE. As shown in Table \ref{table: clip},  MIL-NCE has a lower performance than CLIP, but with the help of our schemata, it achieves comparable performance to CLIP. In addition, the performance of CLIP also increases significantly by using our schemata. This indicates that our schemata are transferable across different video-text models to improve the video retrieval performance.

\section{Conclusion}
We propose a schema induction and generalization system that improves instructional video retrieval performance. We demonstrate that the induced schemata benefit video retrieval on unseen tasks, and our IER system outperforms other methods. 
In the future, we plan to investigate the structure of our schemata, such as the temporal order, and discover other applications of schemata.
{\small
\bibliographystyle{ieee_fullname}
\bibliography{egbib}
}

\end{document}